\documentclass[10pt,twocolumn,letterpaper]{article}

\usepackage{cvpr}              
\usepackage[T1]{fontenc}
\usepackage{graphicx}
\usepackage{hyperref}
\usepackage{xcolor}
\usepackage{float}
\usepackage{pgf-pie}
\usepackage{amsmath}
\usepackage{booktabs}
\usepackage{colortbl}
\usepackage{tikz}
\usepackage{booktabs}
\usepackage{adjustbox}
\usepackage[table]{xcolor}
\usepackage[most]{tcolorbox}
\usepackage{tikz}
\usetikzlibrary{arrows.meta}
\usepackage{enumitem}
\usepackage{subcaption}
\usepackage{pgfplots}

\pgfplotsset{compat=1.17}
\usetikzlibrary{shapes, arrows, positioning}
\def\paperID{*****} 
\def\confName{EIML3@NeurIPS}
\def\confYear{2026}

\title{What Do Medical Vision–Language Models Learn in Radiology? \\ 
Transfer, Alignment, and Source-Proxy Leakage Under Distribution Shift
}

\author{
\textbf{Ayoub Louaye Bouaziz$^{1}$} \quad
\textbf{Lokmane Chebouba$^{2}$ \quad
Yassine Himeur$^{3}$} \\
\vspace{0.3em}
$^{1}$LaTIM, Inserm, University of Western Brittany
, Brest, France \\
$^{2}$University of Constantine, Algeria \\
$^{3}$University of Dubai, UAE \\
\vspace{0.3em}
{\tt\small
\textbf{ayoublouaye.bouaziz@student.umc.edu.dz \quad
lokmane.chebouba@umc.edu.dz \quad
yhimeur@ud.ac.ae}
}
}
\begin{document}
\maketitle
\begin{abstract}
Medical vision--language models (VLMs) can appear reliable in-domain while failing when acquisition domain, paired supervision, or evaluation protocol changes. We study this failure mode as a representation-level blind spot relevant to epistemic intelligence, without claiming a formal estimator of epistemic uncertainty. Using NIH ChestXray14 and CheXpert, we first isolate source-only cross-dataset visual transfer from unsupervised domain-adaptation diagnostics. Using PadChest and OpenI, we then evaluate multimodal alignment under strict pair-index retrieval and quantify metadata-derived source-proxy information retained in frozen embeddings. Self-supervised visual initialization improves NIH-to-CheXpert transfer over supervised ImageNet initialization in matched ResNet-18 comparisons, whereas adversarial adaptation is useful only in a narrow regime and becomes unstable as adversarial pressure increases. Multimodal exact-pair retrieval remains low under external OpenI stress testing, and source-proxy information remains recoverable from learned representations. Qualitative nearest-neighbor and Grad-CAM analyses show clinically plausible cross-dataset structure and thoracic attention patterns in many cases, while device-heavy and false-positive cases remain ambiguous. Auxiliary architecture checks are task-dependent and do not support a universal backbone ranking. Overall, the study shows that apparent competence under a single protocol can conceal transfer, alignment, and shortcut-related failure modes, motivating stress-tested evaluation of medical VLMs under distribution shift.
\end{abstract}

\section{Introduction}
\label{sec:intro}

Deep learning models for chest X-ray analysis can degrade across hospitals because acquisition pipelines, patient populations, and reporting conventions change across institutions. Models may therefore exploit scanner, view, or documentation regularities that do not transfer with pathology~\cite{zech2018confounding,oakdenrayner2020hidden}. This problem becomes more difficult for vision--language models (VLMs), where image and text streams can each carry dataset-specific signals.

Medical VLMs align radiographs with clinical text and can support zero-shot prediction or retrieval~\cite{radford2021clip,zhang2020convirt,huang2021gloria,zhang2022chexzero,you2023cxrclip}. Yet strong in-domain performance does not establish that the learned representation remains useful when the image domain, report style, or supervision regime changes. Recent reviews likewise identify robustness, grounding, and evaluation under dataset variation as open problems for medical vision--language systems~\cite{ryu2025medicalvlmreview}.

This paper asks a narrower question: what failure modes become visible when chest X-ray representations are stress-tested across datasets? This framing is directly relevant to epistemic intelligence under distribution shift. A system may appear competent on familiar data while its representation fails to support the same decision or alignment externally. We do not estimate epistemic uncertainty and do not claim a formal notion of what a model ``knows.'' Instead, we operationalize one EIML-relevant blind spot through controlled tests of transfer, alignment, and recoverable source-proxy information.

We use NIH ChestXray14~\cite{wang2017chestxray14} and CheXpert~\cite{irvin2019chexpert} for visual transfer, and PadChest~\cite{bustos2020padchest} and OpenI~\cite{demner2016openi} for paired multimodal evaluation. The primary tier uses matched ResNet-18 encoders so that initialization effects are not confounded with backbone changes. A separate auxiliary tier probes architecture sensitivity without using it to support causal claims about architecture superiority.

Our prior work studied feature-level site leakage in a single-modal cross-hospital setting~\cite{bouaziz2026feature}. The present study changes the scientific scope in four ways: it evaluates image--text alignment, separates paired-supervision effects from source-only visual transfer, tests external paired retrieval, and measures metadata-derived source-proxy recoverability in multimodal representations. This distinction is important because the present claims concern multimodal evaluation behavior rather than a new site-invariant encoder.

\paragraph{Contributions.}
\begin{itemize}[leftmargin=*]
    \item We present a controlled stress-test of medical VLM representations under distribution shift, separating source-only visual transfer, unsupervised domain-adaptation diagnostics, multimodal pair-index retrieval, and source-proxy leakage analysis.
    \item We show that self-supervised visual initialization improves matched NIH-to-CheXpert transfer over supervised ImageNet initialization, while stronger adversarial adaptation becomes unstable.
    \item We define a dual-scale multimodal \emph{evaluation setting}, not a new benchmark, using PadChest for paired training/in-domain analysis and OpenI for external stress testing.
    \item We make the leakage claim precise: OpenI has no explicit hospital-site labels, so we report \emph{metadata-derived source-proxy leakage} rather than site leakage.
    \item We complement scalar metrics with qualitative cross-dataset retrieval and Grad-CAM analyses, and we treat architecture results as task-dependent sensitivity checks rather than a backbone ranking.
\end{itemize}

\begin{figure*}[t]
\centering
\resizebox{\textwidth}{!}{\input{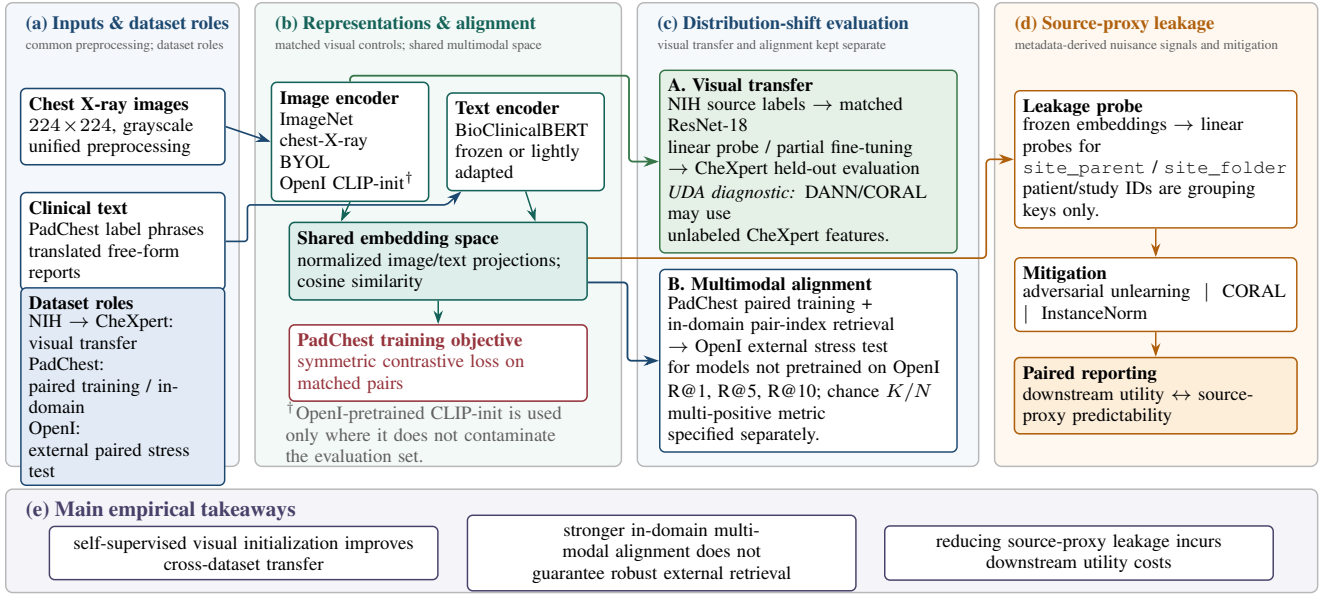}}
\caption{\textbf{Overview of the corrected leakage-aware evaluation framework.} The pipeline separates three questions that must not be conflated: source-only visual transfer, multimodal alignment under an external dataset shift, and recoverability of metadata-derived source proxies. OpenI is considered held out only for models whose pretraining did not use OpenI. The OpenI-pretrained CLIP initialization is therefore excluded from held-out OpenI quantitative claims.}
\label{fig:framework}
\end{figure*}

\section{Related Work}
\label{sec:related}

\subsection{Chest X-ray Generalization and Self-Supervision}
Cross-institution generalization remains difficult in chest radiography because models can exploit acquisition and cohort-specific shortcuts~\cite{zech2018confounding,oakdenrayner2020hidden,badgeley2019deep}. Self-supervised learning can improve medical image transfer when labeled data are limited or shifted~\cite{azizi2021bigsslmedical,taleb2020selfsupervised3d}. Our visual-transfer tier uses this literature as a controlled representation baseline rather than as an architectural contribution.

\subsection{Medical Vision--Language Representation Learning}
ConVIRT, GLoRIA, CheXzero, and CXR-CLIP illustrate the progression from paired medical contrastive pretraining to chest X-ray-specific language--image systems~\cite{zhang2020convirt,huang2021gloria,zhang2022chexzero,you2023cxrclip}. These methods demonstrate useful image--text structure, but evaluation can depend strongly on dataset construction, prompt formulation, and candidate-pool definition. Our focus is therefore diagnostic: we ask which conclusions survive conservative cross-dataset stress tests.

\subsection{Leakage, Grounding, and Evaluation}
Shortcut learning and hidden stratification motivate patient-disjoint splits and explicit nuisance probes~\cite{badgeley2019deep,oakdenrayner2020hidden}. In multimodal data, the problem can be amplified because image and report modalities may share source-specific regularities. We therefore pair utility metrics with source-proxy probes and avoid equating metadata-derived proxies with true hospital-site labels.

\section{Experimental Setup}
\label{sec:setup}

\subsection{Datasets and Roles}
\label{sec:datasets}
We use four public chest X-ray datasets with fixed roles. NIH ChestXray14 is the labeled source for visual transfer and CheXpert is the external target. PadChest is the paired training corpus and in-domain multimodal evaluation set. OpenI is the external paired stress-test set. All images are converted to grayscale, resized to $224\times224$, and normalized under a common pipeline. Patient-disjoint splits are used when patient identifiers are available. Grouping and label-mapping details are provided in the supplement.

\subsection{Primary Matched Visual Encoders}
\label{sec:image_encoders}
The primary controlled tier uses ResNet-18~\cite{he2016resnet} and varies initialization while keeping architecture fixed:
\begin{itemize}[leftmargin=*]
    \item \textbf{ImageNet}: supervised natural-image initialization~\cite{deng2009imagenet}.
    \item \textbf{BYOL}: self-supervised chest X-ray initialization~\cite{grill2020byol}.
    \item \textbf{OpenI CLIP-init}: a CLIP-style image encoder pretrained on OpenI, used only where this does not contaminate the evaluation set.
\end{itemize}
The OpenI-pretrained CLIP initialization is \textbf{not} a held-out OpenI baseline and is excluded from external OpenI quantitative claims.

\subsection{Source-Only Transfer and Unsupervised Adaptation}
\label{sec:transfer_protocol}
For source-only transfer, the classification head and any trainable visual layers use NIH labels only. CheXpert labels are never used for parameter updates, hyperparameter selection, or early stopping. We report a frozen-backbone linear probe and partial fine-tuning of the final residual block plus classifier.

DANN~\cite{ganin2016dann} and CORAL~\cite{sun2016coral} are reported separately as \emph{unsupervised domain-adaptation} diagnostics because they use unlabeled target-domain features to construct domain or covariance losses. They are therefore not presented as pure domain-generalization methods. We use scheduled adversarial strength and monitor the complete training trajectory to expose instability.

\subsection{Multimodal Construction and Retrieval}
\label{sec:alignment_eval}
We pair an image encoder with BioClinicalBERT~\cite{alsentzer2019bioclinicalbert}, project both modalities to a normalized shared space, and optimize symmetric contrastive loss on paired PadChest data. We examine two text constructions: standardized label-derived English phrases and automatically translated free-form reports.

Retrieval is reported as a strict \emph{pair-index retrieval diagnostic} with Recall@$K$, $K\in\{1,5,10\}$. This wording is deliberate. PadChest templates can repeat across samples, and OpenI can associate multiple images with report-level text. Consequently, an index-exact miss is not necessarily a semantic mismatch. We therefore do not interpret these values as a clinical semantic-retrieval benchmark. For OpenI, the processed evaluation index contains $N=6800$ candidate entries. We do not equate this count with the number of unique OpenI reports. Chance is $K/N$. The PadChest diagnostic analogously uses $N=15000$ candidate entries.

The label-derived PadChest templates are used to test sensitivity to controlled linguistic formulation. Because duplicate templates can create multiple valid semantic positives, they are not used to support claims about exact one-to-one semantic retrieval. A duplicate-aware or multi-positive retrieval protocol would be required for that stronger claim.

\subsection{Source-Proxy Leakage}
\label{sec:leakage_defs}
OpenI does not provide explicit hospital-site labels. We therefore use two metadata-derived source proxies available in the processed data: \texttt{site\_parent} and \texttt{site\_folder}. These names are retained to match the implementation, but they are not treated as verified site identities. Linear probes predict the proxy class from frozen embeddings.

The split rule is group-aware: proxy classes may occur in both train and test because they are the prediction targets, while patient or study groups are kept disjoint to prevent memorization of repeated samples. We do not use patient identity as a probe target. Patient/study identifiers are grouping keys only.

\subsection{Leakage Mitigation and Architecture Sensitivity}
\label{sec:leakage_methods}
We evaluate adversarial proxy unlearning, CORAL alignment, and an InstanceNorm ablation~\cite{ulyanov2016instancenorm}. Each method is reported jointly with downstream utility and proxy predictability.

A separate architecture-sensitivity tier uses ResNet-50, DenseNet-121, EfficientNet-B0, ViT-S, Swin-T, and a CLIP-initialized ResNet-50~\cite{he2016resnet,huang2017densenet,tan2019efficientnet,dosovitskiy2021vit,liu2021swin}. These runs are auxiliary. Their task definitions differ from the primary matched transfer experiment, so we do not combine heterogeneous metrics into a single architecture-ranking table.

\subsection{Implementation Notes}
\label{sec:impl}
Experiments are implemented in PyTorch. Adam or AdamW~\cite{kingma2015adam,loshchilov2019adamw} is used depending on the experiment, with fixed settings within each matched comparison and validation-based early stopping where applicable. CheXpert uncertainty handling and the cross-dataset pathology mapping are specified in the supplement. We report variability only for experiments for which repeated-run statistics are available, rather than implying a uniform seed count across all runs. The supplement includes an explicit reproducibility audit that separates archived settings from run-level metadata that were not preserved.

\section{Results}
\label{sec:results}

\subsection{Matched NIH-to-CheXpert Transfer}
\label{sec:results_transfer}
Table~\ref{tab:transfer_main} reports the central quantitative evidence for the matched ResNet-18 comparison. BYOL initialization improves AUC over ImageNet initialization under both reported regimes. The comparison is architecture-matched and uses NIH labels only.

\begin{table}[t]
\centering
\small
\caption{NIH ChestXray14 $\rightarrow$ CheXpert transfer AUC with matched ResNet-18 encoders. The archived manuscript contains point estimates but not the seed-level records required to reconstruct mean$\pm$std; no significance claim is made.}
\label{tab:transfer_main}
\begin{tabular}{lcc}
\toprule
\rowcolor{cvprheader}
\textbf{Initialization} & \textbf{Linear probe} & \textbf{Partial fine-tune} \\
\midrule
\rowcolor{cvprred}
ImageNet & 0.82 & 0.85 \\
\rowcolor{cvprgreen}
BYOL & \textbf{0.87} & \textbf{0.88} \\
\bottomrule
\end{tabular}
\end{table}

The result supports a limited claim: under this matched setup, chest X-ray SSL provides a better starting representation than supervised ImageNet initialization. It does not establish that initialization dominates architecture in general.

\subsection{Unsupervised Domain-Adaptation Behavior}
\label{sec:results_dann}
\begin{figure}[t]
\centering
\includegraphics[width=\linewidth]{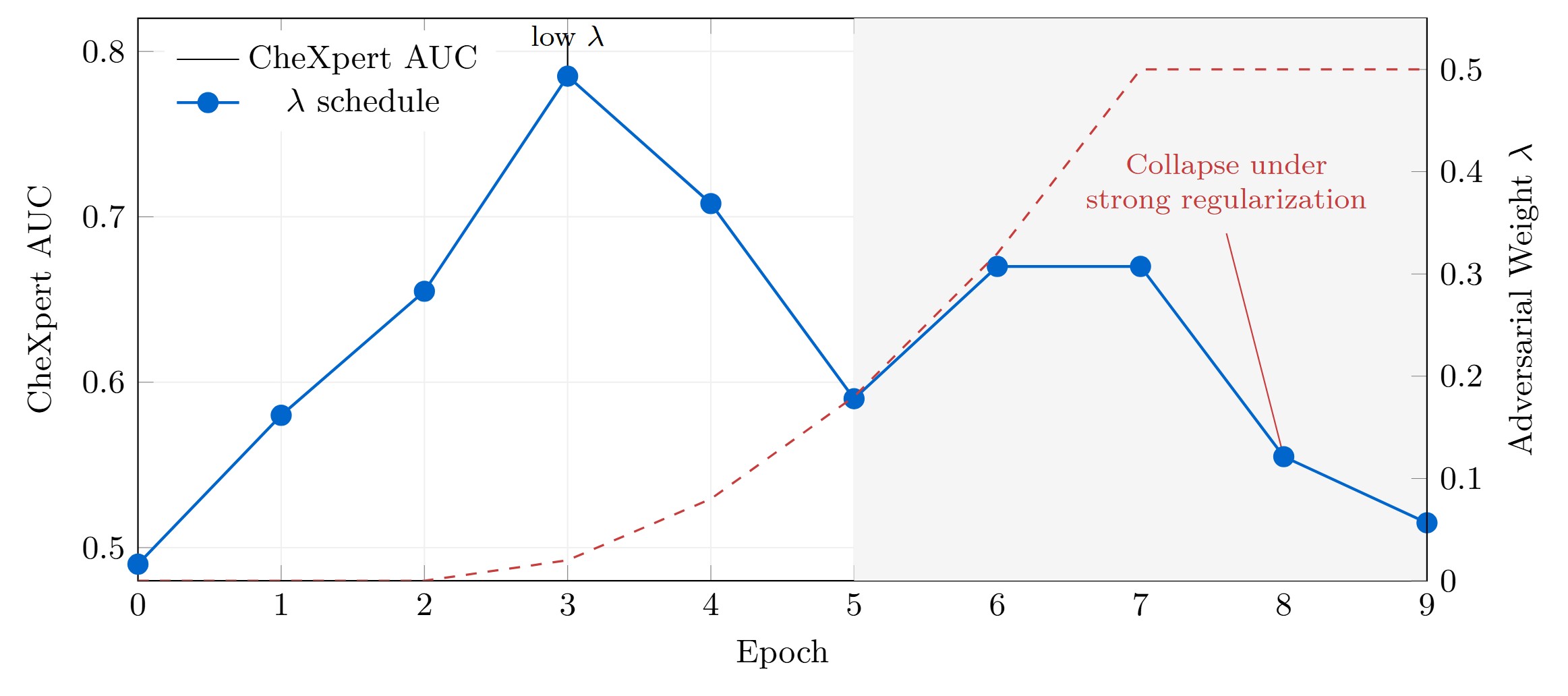}
\caption{\textbf{DANN sensitivity on CheXpert.} Target AUC across training epochs under a scheduled gradient-reversal weight. The trajectory is used as a stability diagnostic, not as evidence for a universal benefit of domain adaptation.}
\label{fig:dann_instability}
\end{figure}

Figure~\ref{fig:dann_instability} shows that DANN behavior is sensitive to adversarial strength. Target AUC rises early and then degrades as adversarial pressure becomes stronger. CORAL is less erratic in the observed runs, but neither method is treated as a substitute for source representation quality.

\subsection{Multimodal Pair-Index Retrieval}
\label{sec:results_vlm}
Table~\ref{tab:retrieval_main} reports the strict pair-index diagnostic. The processed OpenI evaluation index contains 6800 candidate entries. This count is not interpreted as the number of unique reports. The OpenI-pretrained CLIP initialization is omitted from the held-out OpenI rows because it has seen OpenI during pretraining.

\begin{table*}[t]
\centering
\small
\caption{Strict pair-index image-to-text retrieval. OpenI and PadChest pool sizes refer to candidate entries in the processed evaluation index. These numbers should not be interpreted as semantic retrieval accuracy when multiple indices can share equivalent report text. OpenI CLIP-init is intentionally omitted from the held-out OpenI comparison because that initialization was pretrained on OpenI.}
\label{tab:retrieval_main}
\begin{tabular}{llcccc}
\toprule
\rowcolor{cvprheader}
\textbf{Dataset} & \textbf{Initialization} & \textbf{R@1} & \textbf{R@5} & \textbf{R@10} & \textbf{External held-out?} \\
\midrule
\rowcolor{cvprrow}
OpenI, $N=6800$ entries & Chance & 0.000147 & 0.000735 & 0.001471 & -- \\
\rowcolor{blue!6}
OpenI, $N=6800$ entries & ImageNet + Text & 0.0006 & 0.0010 & 0.0020 & Yes \\
\rowcolor{orange!8}
OpenI, $N=6800$ entries & BYOL + Text & 0.0002 & 0.0010 & 0.0022 & Yes \\
\midrule
\rowcolor{cvprrow}
PadChest, $N=15000$ entries & Chance & 0.000067 & 0.000333 & 0.000667 & -- \\
\rowcolor{blue!6}
PadChest, $N=15000$ entries & ImageNet + Text & 0.0015 & 0.0019 & 0.0024 & In-domain \\
\rowcolor{orange!8}
PadChest, $N=15000$ entries & BYOL + Text & 0.0007 & 0.0015 & \textbf{0.0026} & In-domain \\
\rowcolor{green!8}
PadChest, $N=15000$ entries & OpenI CLIP-init + Text & 0.0009 & 0.0014 & 0.0022 & In-domain \\
\bottomrule
\end{tabular}
\end{table*}

The external OpenI numbers are low in absolute terms and remain close to chance at larger $K$. The defensible conclusion is therefore specific: exact pair identity learned from PadChest does not transfer strongly to the processed OpenI pool under this strict diagnostic. Because duplicate or report-level positives are not modeled, we do not generalize this result to all forms of semantic medical image--text retrieval.

\subsection{Qualitative Representation Structure}
\label{sec:results_qualitative}
\begin{figure}[t]
    \centering
    \includegraphics[width=\linewidth]{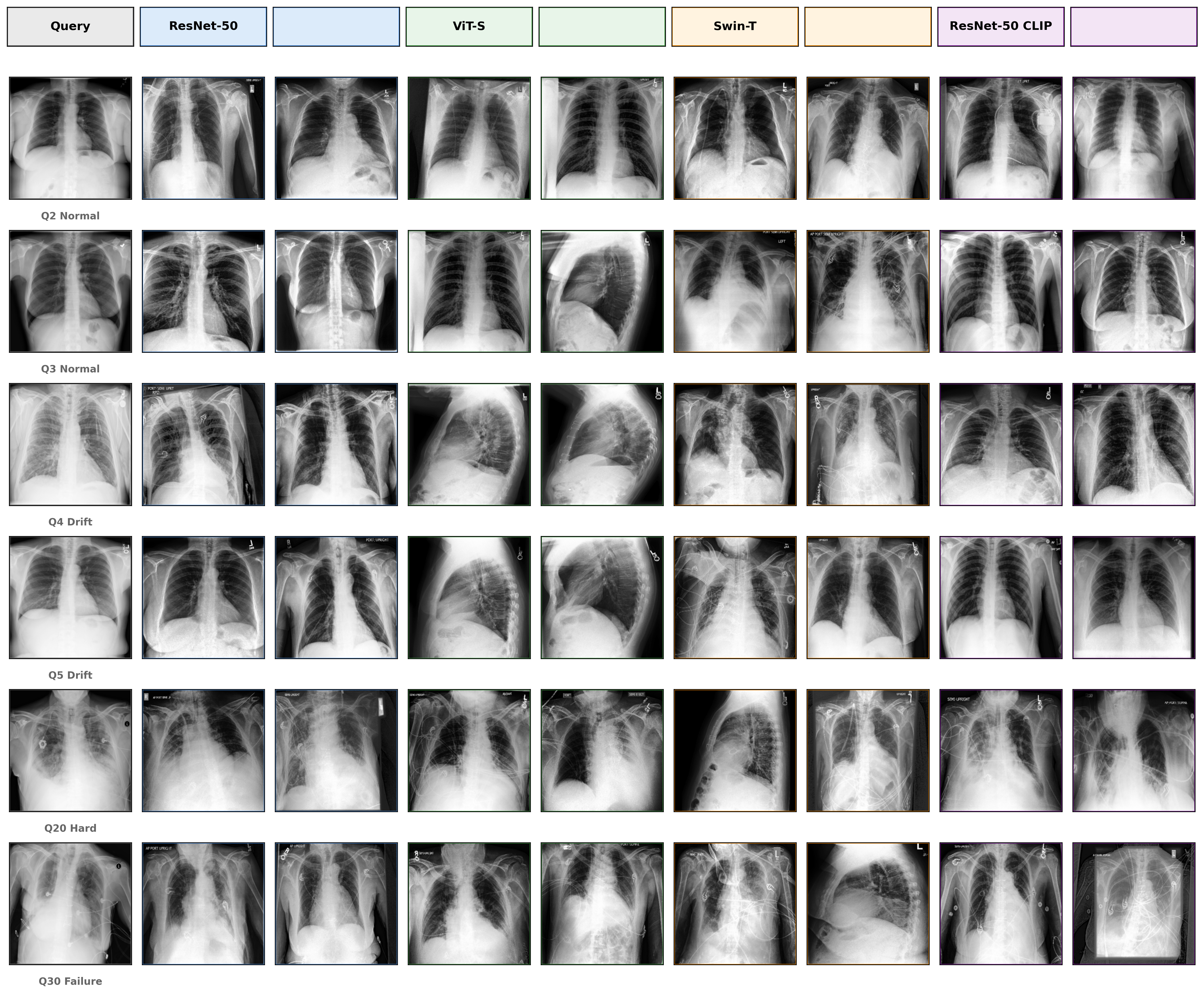}
    \caption{\textbf{Cross-dataset nearest neighbors in frozen image-representation space.} For each NIH query, top CheXpert neighbors are shown for an SSL ResNet-50, ViT-S, Swin-T, and a CLIP-style ResNet-50 image encoder. Clean cases often preserve plausible anatomy/view similarity, while harder cases expose ambiguity and view sensitivity.}
    \label{fig:qual_retrieval}
\end{figure}

\begin{figure}[t]
    \centering
    \includegraphics[width=\linewidth]{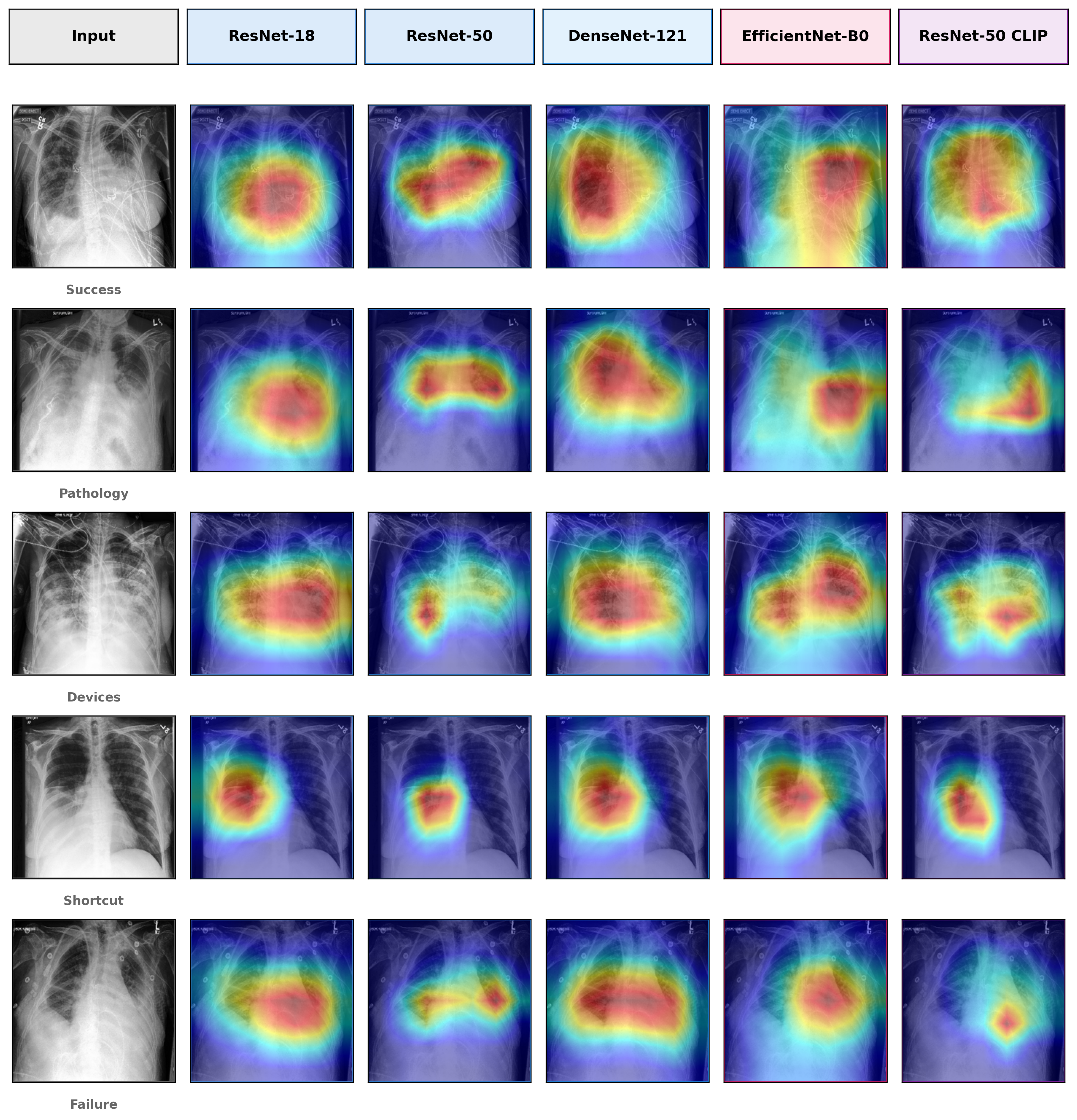}
    \caption{\textbf{Grad-CAM for consolidation classification on representative CheXpert cases.} The maps frequently concentrate attention within thoracic regions, while device-heavy and false-positive cases are more ambiguous. These visualizations are qualitative attention diagnostics, not a localization benchmark.}
    \label{fig:qual_cam}
\end{figure}

The two figures provide evidence about representation behavior that scalar AUC alone cannot show. They support a modest claim that clinically plausible visual structure is often retained. They do not establish lesion localization accuracy or causal grounding.

\subsection{Source-Proxy Leakage and Mitigation}
\label{sec:results_leakage}
The OpenI probes show that metadata-derived source proxies remain recoverable from frozen representations. Because the proxies come from dataset organization metadata rather than verified hospital acquisition labels, the correct interpretation is source-proxy recoverability, not proof of hospital-site identification.

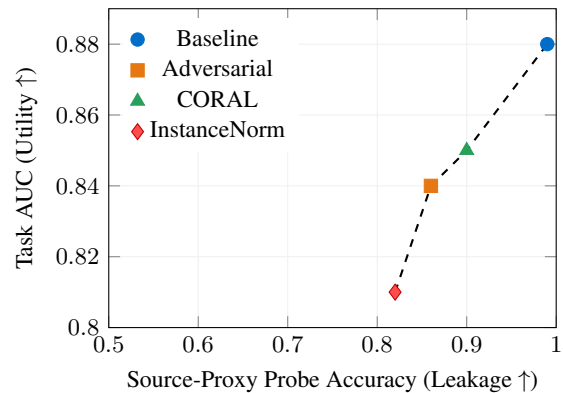
\begin{figure}[t]
\centering
\begin{tikzpicture}
\begin{axis}[
width=0.9\linewidth,
height=5.8cm,
xlabel={Source-Proxy Probe Accuracy (Leakage $\uparrow$)},
ylabel={Task AUC (Utility $\uparrow$)},
xmin=0.50, xmax=1.00,
ymin=0.80, ymax=0.89,
grid=major,
grid style={gray!12},
tick label style={font=\small},
label style={font=\small},
legend style={at={(0.03,0.97)},anchor=north west,draw=none,font=\small},
]
\addplot[only marks,mark=*,mark size=2.5pt,draw=cvprblue2,fill=cvprblue2] coordinates {(0.99,0.88)}; \addlegendentry{Baseline}
\addplot[only marks,mark=square*,mark size=2.5pt,draw=cvprorange2,fill=cvprorange2] coordinates {(0.86,0.84)}; \addlegendentry{Adversarial}
\addplot[only marks,mark=triangle*,mark size=3pt,draw=cvprgreen2,fill=cvprgreen2] coordinates {(0.90,0.85)}; \addlegendentry{CORAL}
\addplot[only marks,mark=diamond*,mark size=3pt,draw=red!75!black,fill=red!70] coordinates {(0.82,0.81)}; \addlegendentry{InstanceNorm}
\addplot[black,dashed,thick] coordinates {(0.82,0.81) (0.86,0.84) (0.90,0.85) (0.99,0.88)};
\end{axis}
\end{tikzpicture}
\caption{\textbf{Utility--source-proxy leakage trade-off.} InstanceNorm gives the largest reduction in proxy accuracy and also the largest utility loss. Adversarial unlearning is intermediate, while CORAL produces a milder change.}
\label{fig:utility_leakage_tradeoff}
\end{figure}

Figure~\ref{fig:utility_leakage_tradeoff} shows that InstanceNorm reduces proxy accuracy most aggressively and incurs the largest utility cost. Adversarial unlearning provides an intermediate reduction with a smaller, but still visible, utility loss. This result supports a trade-off claim only for the evaluated proxy and task.

\subsection{Architecture Sensitivity Is Task-Dependent}
\label{sec:results_architecture}
Table~\ref{tab:arch_sensitivity} reports a single auxiliary consolidation task, separate from the different CheXpert-wide protocol. Swin-T reaches 0.5373 in this run and is marked unstable, consistent with the supplementary table.

\begin{table}[t]
\centering
\small
\caption{Auxiliary consolidation-classifier AUC. These results are a sensitivity check and are not directly comparable with the primary NIH-to-CheXpert transfer table.}
\label{tab:arch_sensitivity}
\begin{tabular}{lcc}
\toprule
\rowcolor{cvprheader}
\textbf{Backbone} & \textbf{AUC} & \textbf{Note} \\
\midrule
\rowcolor{cvprgreen}
ResNet-18 & \textbf{0.9257} & Stable \\
\rowcolor{cvprrow}
ResNet-50 & 0.9163 & Stable \\
DenseNet-121 & 0.9200 & Stable \\
\rowcolor{cvprrow}
EfficientNet-B0 & 0.9203 & Stable \\
ResNet-50 CLIP & 0.9155 & Stable \\
\rowcolor{orange!8}
ViT-S & 0.8774 & Lower on this task \\
\rowcolor{cvprred}
Swin-T & 0.5373 & Unstable exploratory \\
\bottomrule
\end{tabular}
\end{table}

These results do not justify the previous statement that architecture differences are universally modest. The defensible interpretation is task-dependent. CNN-family encoders cluster closely on this consolidation check, ViT-S is lower, and Swin-T is unstable. A separate CheXpert-only architecture table in the supplement uses a different label space and should be read only as an auxiliary evaluation.

\section{Limitations}
\label{sec:limitations}
First, the multimodal study uses one large paired training corpus and one external paired stress-test dataset. Second, the current retrieval archive supports exact pair-index Recall@$K$ but does not preserve a validated equivalence map for all duplicate or report-level positives. We therefore define a duplicate-aware multi-positive metric in the supplement but do not fabricate an uncomputed score; the reported exact-index metric remains a conservative stress-test diagnostic. Third, the central matched transfer values are archived as point estimates, while the seed-level records required to reconstruct mean$\pm$std are unavailable in the supplied experiment archive. We consequently make no significance claim for the 0.82--0.88 differences. Fourth, the source bundle does not preserve a complete run-level hyperparameter ledger for every experimental family, so the supplement explicitly marks which settings are known and which exact values are unavailable. Fifth, the exact GPT-4 API/model snapshot used for PadChest translation was not preserved; temperature zero reduces decoding variability but does not guarantee identical regeneration. Beyond these four reproducibility limitations, OpenI provides no verified hospital-site labels, the architecture sweep is auxiliary and heterogeneous across tasks, the study is restricted to chest radiography, and we do not measure calibrated epistemic uncertainty, abstention, or formal unknown-unknown detection.

\section{Conclusion}
\label{sec:conclusion}
We studied chest X-ray VLMs through controlled stress tests of representation transfer, multimodal pair-index alignment, and metadata-derived source-proxy recoverability. In matched NIH-to-CheXpert comparisons, BYOL initialization improves over ImageNet initialization. Unsupervised adversarial adaptation is useful only in a narrow regime and becomes unstable as adversarial pressure increases. Under external OpenI evaluation, strict exact-pair retrieval remains weak, while source-proxy signals remain recoverable from frozen embeddings. Qualitative retrieval and Grad-CAM results show plausible cross-dataset structure and thoracic attention patterns but also persistent ambiguity in difficult cases. Auxiliary architecture checks are task-dependent and do not support a universal ranking. These findings fit an epistemic-intelligence perspective as an evaluation result: apparent in-domain competence can coexist with representation-level blind spots under distribution shift. The paper therefore argues for explicit stress tests, contamination-aware baselines, and precise claims about what evaluation protocols actually measure.

{
    \footnotesize
    \setlength{\bibsep}{0pt}
    \bibliographystyle{ieeenat_fullname}
    \bibliography{main}

@String(CVPR= {IEEE Conf. Comput. Vis. Pattern Recog.})

@String(ICCV= {Int. Conf. Comput. Vis.})

@String(ECCV= {Eur. Conf. Comput. Vis.})

@String(ICLR = {Int. Conf. Learn. Represent.})

@String(AAAI = {AAAI})

@inproceedings{wang2017chestxray14,
title={ChestX-ray8: Hospital-scale Chest X-ray Database and Benchmarks on Weakly-Supervised Classification},
author={Wang, Xiaosong and Peng, Yifan and Lu, Le and Lu, Zhiyong and Bagheri, Mohammadhadi and Summers, Ronald},
booktitle={CVPR},
year={2017}
}

@inproceedings{irvin2019chexpert,
title={CheXpert: A Large Chest Radiograph Dataset with Uncertainty Labels and Expert Comparison},
author={Irvin, Jeremy and Rajpurkar, Pranav and Ko, Michael and Yu, Yifan and Ciurea-Ilcus, Silviu and Chute, Christopher and Marklund, Henrik and Haghgoo, Babak and Ball, Robyn and Shpanskaya, Katie and others},
booktitle={AAAI},
year={2019}
}

@article{bustos2020padchest,
title={PadChest: A Large Chest X-ray Image Dataset with Multi-label Annotated Reports},
author={Bustos, Aurelia and Pertusa, Antonio and Salinas, Jose-Maria and de la Iglesia-Vayá, María},
journal={Scientific Data},
year={2020}
}

@article{demner2016openi,
title={Preparing a collection of radiology examinations for distribution and retrieval},
author={Demner-Fushman, Dina and Kohli, Marc and Rosenman, Marc and Shooshan, Sonya and Rodriguez, Laritza and Antani, Sameer and Thoma, George and McDonald, Clement},
journal={Journal of the American Medical Informatics Association},
year={2016}
}

@inproceedings{radford2021clip,
title={Learning Transferable Visual Models From Natural Language Supervision},
author={Radford, Alec and others},
booktitle={ICML},
year={2021}
}

@article{zhang2022chexzero,
title={CheXzero: Zero-shot Chest X-ray Classification via Contrastive Language–Image Pretraining},
author={Zhang, Yuhao and others},
journal={Nature Biomedical Engineering},
year={2022}
}

@article{zech2018confounding,
title={Variable generalization performance of a deep learning model to detect pneumonia in chest radiographs},
author={Zech, John and Badgeley, Marcus and Liu, Manway and Costa, Andrew and Titano, Joshua and Oermann, Eric},
journal={PLoS Medicine},
year={2018}
}

@article{oakdenrayner2020hidden,
title={Hidden stratification causes clinically meaningful failures in machine learning for medical imaging},
author={Oakden-Rayner, Luke and Dunnmon, Jared and Carneiro, Gustavo and Re, Christopher},
journal={Nature Medicine},
year={2020}
}

@inproceedings{azizi2021bigsslmedical,
title={Big Self-Supervised Models Advance Medical Image Classification},
author={Azizi, Shekoofeh and Mustafa, Basil and Ryan, Fiona and others},
booktitle={ICCV},
year={2021}
}

@article{taleb2020selfsupervised3d,
title={3D self-supervised methods for medical imaging},
author={Taleb, A. and others},
journal={Medical Image Analysis},
year={2020}
}

@inproceedings{huang2021gloria,
title={GLoRIA: A Multimodal Global-Local Representation Learning Framework for Label-efficient Medical Image Recognition},
author={Huang, Xinyue and others},
booktitle={ICCV},
year={2021}
}

@article{zhang2020convirt,
title={ConVIRT: Contrastive Learning of Medical Visual Representations from Paired Images and Text},
author={Zhang, Yuhao and others},
journal={arXiv preprint arXiv:2010.00747},
year={2020}
}

@article{badgeley2019deep,
title={Deep learning predicts hip fracture using confounding patient and healthcare variables},
author={Badgeley, Marcus and others},
journal={npj Digital Medicine},
year={2019}
}

@inproceedings{he2016resnet,
title={Deep Residual Learning for Image Recognition},
author={He, Kaiming and Zhang, Xiangyu and Ren, Shaoqing and Sun, Jian},
booktitle={CVPR},
year={2016}
}

@inproceedings{deng2009imagenet,
title={ImageNet: A Large-Scale Hierarchical Image Database},
author={Deng, Jia and others},
booktitle={CVPR},
year={2009}
}

@inproceedings{grill2020byol,
title={Bootstrap Your Own Latent: A New Approach to Self-Supervised Learning},
author={Grill, Jean-Bastien and others},
booktitle={NeurIPS},
year={2020}
}

@inproceedings{ganin2016dann,
title={Domain-Adversarial Training of Neural Networks},
author={Ganin, Yaroslav and others},
booktitle={JMLR},
year={2016}
}

@inproceedings{sun2016coral,
title={Deep CORAL: Correlation Alignment for Deep Domain Adaptation},
author={Sun, Baochen and Saenko, Kate},
booktitle={ECCV Workshops},
year={2016}
}

@article{alsentzer2019bioclinicalbert,
title={Publicly Available Clinical BERT Embeddings},
author={Alsentzer, Emily and Murphy, John and Boag, Willie and Weng, Wei-Hung and Jin, Di and Naumann, Tristan and McDermott, Matthew},
journal={NAACL Clinical NLP Workshop},
year={2019}
}

@inproceedings{kingma2015adam,
title={Adam: A Method for Stochastic Optimization},
author={Kingma, Diederik and Ba, Jimmy},
booktitle={ICLR},
year={2015}
}

@article{loshchilov2019adamw,
title={Decoupled Weight Decay Regularization},
author={Loshchilov, Ilya and Hutter, Frank},
journal={ICLR},
year={2019}
}

@inproceedings{ulyanov2016instancenorm,
title={Instance Normalization: The Missing Ingredient for Fast Stylization},
author={Ulyanov, Dmitry and Vedaldi, Andrea and Lempitsky, Victor},
booktitle={arXiv},
year={2016}
}

@inproceedings{liu2021swin,
title={Swin Transformer: Hierarchical Vision Transformer using Shifted Windows},
author={Liu, Ze and others},
booktitle={ICCV},
year={2021}
}

@inproceedings{tan2019efficientnet,
title={EfficientNet: Rethinking Model Scaling for Convolutional Neural Networks},
author={Tan, Mingxing and Le, Quoc},
booktitle={ICML},
year={2019}
}

@inproceedings{huang2017densenet,
title={Densely Connected Convolutional Networks},
author={Huang, Gao and others},
booktitle={CVPR},
year={2017}
}

@inproceedings{dosovitskiy2021vit,
title={An Image is Worth 16x16 Words},
author={Dosovitskiy, Alexey and others},
booktitle={ICLR},
year={2021}
}

@misc{bouaziz2026feature,
  title={Feature-level Site Leakage Reduction for Cross-Hospital Chest X-ray Transfer via Self-Supervised Learning},
  author={Ayoub Louaye Bouaziz and Lokmane Chebouba},
  year={2026},
  eprint={2604.00263},
  archivePrefix={arXiv},
  primaryClass={cs.CV}
}

@inproceedings{you2023cxrclip,
  title={CXR-CLIP: Toward Large Scale Chest X-ray Language-Image Pre-training},
  author={You, Kihyun and Gu, Jawook and Ham, Jiyeon and Park, Beomhee and Kim, Jiho and Hong, Eun Kyoung and Baek, Woonhyuk and Roh, Byungseok},
  booktitle={Medical Image Computing and Computer Assisted Intervention (MICCAI)},
  year={2023}
}

@article{ryu2025medicalvlmreview,
  title={Vision-language foundation models for medical imaging: a review of current practices and innovations},
  author={Ryu, Ji Seung and Kang, Hyunyoung and Chu, Yuseong and Yang, Sejung},
  journal={Biomedical Engineering Letters},
  volume={15},
  pages={809--830},
  year={2025}
}
}

\clearpage

\clearpage
\setcounter{page}{1}
\maketitlesupplementary
\appendix

\section{External Zero-Shot Contextual Baselines}
\label{app:zeroshot_context}

This appendix reports the external zero-shot baselines omitted from the main paper for clarity. These models are included only as contextual reference points under the same unified preprocessing, label mapping, and AUROC-based evaluation protocol. Because they are not architecture-matched controls, they are not used as primary evidence for the paper's controlled conclusions.

\begin{table}[H]
\centering
\small
\caption{Zero-shot contextual baselines on CheXpert under the unified evaluation protocol. We report mean AUC over the 5-label setting for calibration only. These results are not architecture-matched to the primary controlled experiments. All models are evaluated without retraining under the same preprocessing and label-mapping pipeline.}
\label{tab:zeroshot_contextual}
\begin{adjustbox}{width=\columnwidth}
\begin{tabular}{lc}
\toprule
\rowcolor{cvprheader}
Model & Mean AUC (macro, 5-label) \\
\midrule
\rowcolor{cvprrow}
CheXzero & $0.56\ {\color{red}{\pm 0.021}}$ \\
CXR-CLIP & $0.58\ {\color{red}{\pm 0.0175}}$ \\
\bottomrule
\end{tabular}
\end{adjustbox}
\end{table}
The zero-shot results are reported only to contextualize the numerical scale of the primary experiments. Pretraining-data provenance differs across external checkpoints. In particular, public CXR-CLIP checkpoints exist with training compositions that include CheXpert, so the value shown here must not be interpreted as held-out CheXpert generalization unless the exact checkpoint provenance is verified. These baselines are not used for architecture or transfer claims.

\section{Extended Architecture Tables and Exploratory Results}
\label{app:arch_extended}

This appendix reports auxiliary architecture-family comparisons. These runs use task definitions that differ from the primary NIH-to-CheXpert matched transfer experiment and therefore must not be combined into a single architecture-ranking claim. The wide per-pathology table below is a CheXpert-only auxiliary evaluation over the native CheXpert observation set, not the seven-label cross-dataset transfer task defined later in the appendix.

\begin{table*}[!t]
\centering
\small
\caption{CheXpert-only auxiliary per-pathology AUROC over the broader CheXpert observation set. This table is not the NIH-to-CheXpert seven-label transfer experiment and is not directly comparable with the primary matched transfer results. Best result per pathology is highlighted in bold.}
\label{tab:chexpert_full_appendix}
\begin{adjustbox}{width=\textwidth}
\begin{tabular}{lcccccccccccccc|c}
\toprule
\rowcolor{cvprheader}
Model & NF & ECM & Cardio & Opac & Les & Edema & Cons & Pneum & Atel & PTX & Eff & PlO & Frac & Supp & Mean \\
\midrule

\multicolumn{16}{l}{\cellcolor{cvprsection}\textit{CNN backbones}} \\
\midrule
\rowcolor{cvprrow}
ResNet50 & 0.513 & 0.500 & 0.479 & 0.499 & 0.542 & 0.500 & 0.511 & 0.520 & 0.494 & 0.496 & \textbf{0.564} & 0.531 & 0.495 & 0.527 & 0.512 \\
ResNet18 & 0.544 & 0.493 & 0.508 & 0.480 & 0.512 & 0.412 & 0.518 & 0.541 & 0.484 & 0.514 & 0.537 & 0.558 & \textbf{0.560} & 0.510 & \textbf{0.514} \\
\rowcolor{cvprrow}
DenseNet121 & 0.567 & 0.499 & 0.479 & 0.512 & 0.511 & 0.424 & 0.514 & 0.508 & 0.460 & 0.516 & 0.503 & 0.529 & 0.526 & \textbf{0.606} & 0.511 \\
ResNet50d & 0.475 & 0.501 & 0.499 & 0.514 & \textbf{0.579} & 0.422 & 0.502 & 0.507 & \textbf{0.539} & 0.519 & 0.464 & 0.551 & 0.508 & 0.503 & 0.506 \\
\rowcolor{cvprrow}
EfficientNet-V2-S & 0.469 & 0.495 & 0.495 & 0.475 & 0.476 & 0.482 & 0.466 & 0.528 & 0.474 & \textbf{0.532} & 0.488 & 0.533 & 0.549 & 0.578 & 0.503 \\
EfficientNet-B3 & 0.455 & 0.499 & 0.508 & 0.517 & 0.527 & 0.482 & 0.465 & 0.501 & 0.485 & 0.521 & 0.541 & 0.450 & 0.476 & 0.605 & 0.502 \\
\rowcolor{cvprrow}
EfficientNet-B0 & 0.409 & 0.493 & 0.503 & 0.482 & 0.508 & 0.499 & \textbf{0.526} & 0.497 & 0.495 & 0.544 & 0.523 & 0.469 & 0.505 & 0.500 & 0.497 \\
ResNet101 & 0.496 & \textbf{0.508} & \textbf{0.513} & \textbf{0.538} & 0.522 & \textbf{0.505} & 0.492 & \textbf{0.528} & 0.460 & 0.469 & 0.494 & 0.453 & 0.496 & 0.520 & 0.500 \\

\midrule
\multicolumn{16}{l}{\cellcolor{cvprsection}\textit{Modern architectures}} \\
\midrule
\rowcolor{cvprrow}
ViT-Base & 0.478 & 0.492 & 0.505 & 0.505 & 0.534 & \textbf{0.517} & 0.517 & 0.530 & 0.513 & 0.506 & 0.475 & 0.481 & 0.519 & 0.535 & 0.508 \\
ConvNeXt-Base & 0.445 & 0.491 & 0.510 & 0.474 & 0.468 & 0.482 & 0.433 & 0.488 & 0.491 & 0.512 & 0.473 & 0.436 & 0.520 & 0.547 & 0.484 \\
\rowcolor{cvprrow}
Swin-Base & 0.496 & 0.505 & 0.495 & 0.524 & 0.432 & 0.470 & 0.479 & 0.472 & 0.485 & 0.459 & 0.500 & \textbf{0.595} & 0.466 & 0.473 & 0.489 \\

\midrule
\multicolumn{16}{l}{\cellcolor{cvprsection}\textit{Self-supervised representations}} \\
\midrule
\rowcolor{cvprrow}
DINO-ViT & \textbf{0.676} & \textbf{0.519} & 0.512 & 0.452 & 0.480 & 0.444 & 0.479 & 0.509 & 0.520 & 0.480 & 0.471 & 0.443 & 0.515 & 0.474 & 0.498 \\
MAE-ViT & 0.419 & 0.494 & 0.501 & 0.468 & 0.533 & 0.435 & 0.518 & 0.518 & 0.486 & 0.437 & 0.512 & 0.440 & 0.502 & 0.496 & 0.483 \\

\bottomrule
\end{tabular}
\end{adjustbox}
\end{table*}

\begin{table}[!t]
\centering
\small
\caption{Auxiliary consolidation classifier check for the expanded backbone tier. Entries marked unstable are reported for completeness only and are not used as primary qualitative evidence in the main paper.}
\label{tab:arch_aux_appendix}
\begin{adjustbox}{width=\columnwidth}
\begin{tabular}{lcc}
\toprule
\rowcolor{cvprheader}
Backbone & Consolidation AUC & Note \\
\midrule
\rowcolor{cvprrow}
ResNet-18 & 0.9257 & Stable \\
ResNet-50 & 0.9163 & Stable \\
DenseNet-121 & 0.9200 & Stable \\
EfficientNet-B0 & 0.9203 & Stable \\
ResNet-50 CLIP & 0.9155 & Stable \\
ViT-S & 0.8774 & Weaker than CNN family \\
Swin-T & 0.5373 & Unstable exploratory result \\
\bottomrule
\end{tabular}
\end{adjustbox}
\end{table}

The consolidation check is a separate auxiliary task. It should not be numerically combined with the CheXpert-only table above or with the primary NIH-to-CheXpert transfer experiment. Swin-T reached 0.5373 in this auxiliary run and was unstable, so it is reported for completeness rather than used as primary evidence.

\section{Leakage-Aware Evaluation Protocol}

\subsection{Protocol Overview}

\begin{tcolorbox}[
colback=blue!4,
colframe=blue!65!black,
boxrule=0.8pt,
arc=2pt,
title=Leakage-Aware Evaluation Protocol (Summary),
fonttitle=\bfseries,
left=6pt,right=6pt,top=6pt,bottom=6pt
]
This appendix provides the full specification of the leakage-aware evaluation protocol used in this study. The protocol standardizes dataset roles, preprocessing, split construction, evaluation metrics, and diagnostic probes in order to isolate representation transfer, multimodal alignment quality, and potential shortcut signals.

The protocol consists of three complementary experimental settings:

\textbf{(1) Source-only cross-dataset transfer.}  
Visual representations are trained with NIH ChestXray14 labels and evaluated on CheXpert. CheXpert labels are not used for model updates, selection, or early stopping.

\textbf{(2) Multimodal pair-index retrieval.}  
Image--text encoders are trained on paired PadChest data. PadChest is used for in-domain analysis and OpenI (IU X-ray) is used as an external pair-index stress test. OpenI-pretrained CLIP initialization is excluded from held-out OpenI claims.

\textbf{(3) Source-proxy diagnostics.}  
Frozen embeddings are probed for metadata-derived OpenI source proxies. Patient/study identifiers are grouping keys, not probe targets.

Across all experiments the protocol enforces:

\begin{itemize}
\item \textbf{Patient-disjoint splits} whenever patient identifiers are available.
\item \textbf{Unified preprocessing} across datasets.
\item \textbf{Controlled architectures and training regimes}.
\item \textbf{Explicit chance baselines} for retrieval tasks: $R@K = \frac{K}{N}$.
\item \textbf{Paired reporting of utility and leakage metrics}.
\end{itemize}
\end{tcolorbox}

Table~\ref{tab:protocol_datasets} summarizes dataset roles.

\begin{table}[!t]
\centering
\small
\caption{Dataset roles in the leakage-aware evaluation protocol.}
\label{tab:protocol_datasets}
\begin{adjustbox}{width=\columnwidth}
\begin{tabular}{lcc}
\toprule
\rowcolor{cvprheader}
Dataset & Modality & Role in Protocol \\
\midrule
\rowcolor{cvprrow}
NIH ChestXray14 & Image + labels & Source dataset for transfer \\
CheXpert & Image + labels & Target dataset for transfer \\
\rowcolor{cvprrow}
PadChest & Image + reports & Multimodal training + in-domain retrieval \\
OpenI (IU X-ray) & Image + reports & External retrieval evaluation \\
\bottomrule
\end{tabular}
\end{adjustbox}
\end{table}

\subsection{Unified Preprocessing}

To minimize dataset-specific artifacts and ensure fair cross-dataset comparison, all images are processed using a unified preprocessing pipeline before training and evaluation. The same preprocessing steps are applied across NIH ChestXray14, CheXpert, PadChest, and OpenI datasets.

\paragraph{Image normalization pipeline.}
All radiographs are converted to grayscale and resized to a fixed spatial resolution of $224\times224$ pixels using bilinear interpolation. Pixel intensities are normalized to the range $[0,1]$ and subsequently standardized using dataset-independent statistics. No dataset-specific histogram equalization or contrast normalization is applied.

\paragraph{Color channel handling.}
Because chest radiographs are inherently grayscale images, single-channel images are replicated across three channels when required by architectures initialized with ImageNet weights.

\paragraph{Spatial processing.}
Images are resized directly to $224\times224$ without additional cropping or padding in order to preserve anatomical coverage. No center cropping or aspect-ratio preserving padding is applied.

\paragraph{Augmentation policy.}
During training, lightweight data augmentation is applied to improve robustness. This includes random horizontal flipping and small affine transformations (translation and scaling). No dataset-specific augmentations are used. Augmentations are disabled during validation and testing.

\begin{tcolorbox}[
colback=gray!4,
colframe=black!60,
boxrule=0.8pt,
arc=2pt,
title=Preprocessing Pipeline,
fonttitle=\bfseries,
left=6pt,right=6pt,top=6pt,bottom=6pt
]
For each image $x$:
\begin{enumerate}
\item Load image from dataset.
\item Convert to grayscale if necessary.
\item Resize to $224\times224$ using bilinear interpolation.
\item Normalize pixel intensities to $[0,1]$.
\item Replicate channel to three channels if required by the encoder.
\item Apply lightweight augmentation during training only.
\end{enumerate}
\end{tcolorbox}

This unified preprocessing pipeline ensures that performance differences across datasets and models arise from representation quality and training regime rather than differences in image formatting or resolution.

\subsection{Split Construction and Grouping Rules}

To prevent information leakage and ensure realistic evaluation conditions, all experiments use grouped data splits that enforce disjointness at the patient level whenever patient identifiers are available. This ensures that images from the same patient do not appear in both training and evaluation sets.

\paragraph{Patient-disjoint splits.}
For datasets that provide patient identifiers (NIH ChestXray14, CheXpert, and PadChest), splits are constructed such that all studies from a given patient are assigned to a single partition. This prevents models from exploiting patient-specific visual characteristics that could artificially inflate performance.

\paragraph{Cross-dataset transfer setting.}
In the single-modal transfer experiments, models are trained on NIH ChestXray14 and evaluated on CheXpert without any shared patients or studies between datasets. Because the datasets originate from different institutions, cross-dataset transfer provides a realistic domain shift scenario.

\paragraph{Multimodal training and evaluation splits.}
For multimodal experiments, PadChest is used for both training and in-domain retrieval evaluation. The dataset is partitioned into training, validation, and test sets using patient-disjoint splits. OpenI (IU X-ray) is used as an external dataset for cross-dataset retrieval evaluation.

\paragraph{External retrieval dataset (OpenI).}
The OpenI dataset contains image--report pairs collected from a different institution. When patient identifiers are available, patient-disjoint evaluation is enforced. Otherwise, study-level disjointness is maintained to prevent identical image--report pairs from appearing in multiple evaluation subsets.

\paragraph{Grouped splits for source-proxy probes.}
For source-proxy diagnostics, the proxy classes being predicted may occur in both probe-training and probe-test partitions. Disjointness is instead enforced over patient or study groups, so repeated samples from the same patient/study cannot appear in both partitions. This preserves a valid supervised probe while limiting sample-level memorization.

Table~\ref{tab:split_rules} summarizes the grouping keys and split policies used for each dataset.

\begin{table}[!t]
\centering
\small
\caption{Dataset grouping keys and split policies used in the leakage-aware evaluation protocol.}
\label{tab:split_rules}
\begin{adjustbox}{width=\columnwidth}
\begin{tabular}{lccc}
\toprule
\rowcolor{cvprheader}
Dataset & Group Key & Split Type & Usage \\
\midrule
\rowcolor{cvprrow}
NIH ChestXray14 & Patient ID & Patient-disjoint & Training (transfer source) \\
CheXpert & Patient ID & Patient-disjoint & Evaluation (transfer target) \\
\rowcolor{cvprrow}
PadChest & Patient ID & Patient-disjoint & Multimodal training + in-domain retrieval \\
OpenI (IU X-ray) & Patient / Study ID & Study-disjoint & External retrieval evaluation \\
\bottomrule
\end{tabular}
\end{adjustbox}
\end{table}

These split construction rules ensure that performance differences reflect representation quality and cross-domain generalization rather than overlap between training and evaluation data.

\subsection{Label Mapping and Uncertainty Handling}

To enable cross-dataset transfer evaluation between NIH ChestXray14 and CheXpert, we construct a consistent label space across the two datasets. Because the two datasets differ in annotation taxonomy and label extraction pipelines, we define an explicit mapping between overlapping pathology categories.

\paragraph{Shared pathology set.}
Only pathologies that appear in both datasets are used in the transfer experiments. Labels that do not have a clear correspondence between datasets are excluded from the evaluation.

\paragraph{CheXpert uncertainty labels.}
CheXpert annotations include an explicit uncertainty label ($U$) produced by the automatic report labeler. Following common practice in chest X-ray classification benchmarks, uncertainty labels are handled using a fixed policy applied consistently across all experiments.

In our experiments, uncertainty labels are treated as \textbf{negative labels (U-Zeros)}. That is, uncertain observations are mapped to the negative class during training and evaluation. This strategy avoids introducing additional label noise while maintaining consistency with several prior studies.

\paragraph{Label filtering.}
Images that contain missing labels for the selected pathology set are excluded from the corresponding evaluation task. All remaining labels are treated as binary classification targets.

Table~\ref{tab:label_mapping} shows the mapping between NIH ChestXray14 labels and CheXpert observations used in the transfer experiments.

\begin{table}[!t]
\centering
\small
\caption{Mapping between NIH ChestXray14 and CheXpert labels used for cross-dataset transfer evaluation. Only pathologies with clear correspondence across datasets are retained.}
\label{tab:label_mapping}
\begin{adjustbox}{width=\columnwidth}
\begin{tabular}{ll}
\toprule
\rowcolor{cvprheader}
NIH ChestXray14 Label & CheXpert Observation \\
\midrule
\rowcolor{cvprrow}
Atelectasis & Atelectasis \\
Cardiomegaly & Cardiomegaly \\
\rowcolor{cvprrow}
Consolidation & Consolidation \\
Edema & Edema \\
\rowcolor{cvprrow}
Pleural Effusion & Pleural Effusion \\
Pneumonia & Pneumonia \\
\rowcolor{cvprrow}
Pneumothorax & Pneumothorax \\
\bottomrule
\end{tabular}
\end{adjustbox}
\end{table}

This mapping ensures that transfer evaluation reflects differences in learned representations rather than inconsistencies in dataset annotation schemas.

\subsection{Training Regimes}

The primary transfer table reports two regimes whose semantics are explicit in the manuscript.

\paragraph{Linear probing.}
The backbone is frozen and only a linear classification head is optimized using NIH source labels.

\paragraph{Partial fine-tuning.}
The final residual block (layer4 for ResNet-18) and the classifier head are optimized using NIH source labels, while earlier layers remain frozen.

\paragraph{Target-label isolation.}
CheXpert labels are never used for parameter updates, validation selection, or early stopping in the source-only transfer experiment. When DANN or CORAL is enabled, unlabeled CheXpert features may enter the adaptation loss, which is why those runs are reported separately as unsupervised domain adaptation.

\paragraph{Optimization.}
Adam/AdamW-based optimization and validation-based early stopping are used depending on the experiment. Hyperparameters are held fixed within each matched comparison. 

\subsection{Multimodal Alignment and Retrieval Protocol}

Multimodal alignment is evaluated using a strict pair-index image--text retrieval diagnostic. Each query has one designated positive index in the processed evaluation table. This is intentionally not described as a semantic one-to-one benchmark because multiple indices can share equivalent text or report-level content.

Image and text embeddings are produced using the visual encoder and the BioClinicalBERT text encoder, respectively. Both embeddings are projected into a shared latent space and normalized prior to similarity computation.

Retrieval performance is measured using Recall@$K$ (R@$K$), which evaluates whether the designated positive index appears among the top-$K$ candidates. Explicit chance baselines are reported as $K/N$.
\newpage
\begin{tcolorbox}[
enhanced,
breakable,
colback=gray!2,
colframe=black!65,
boxrule=0.9pt,
arc=2pt,
title={Algorithm 1: Multimodal Retrieval Evaluation Protocol},
fonttitle=\bfseries,
left=6pt,right=6pt,top=6pt,bottom=6pt
]

\begin{algorithmic}[1]
\STATE \textbf{Input:} trained image encoder $f_{\theta}$, text encoder $g_{\phi}$, evaluation dataset $\mathcal{D}$ with image--report pairs
\STATE Encode images: $z_i = f_{\theta}(x_i)$
\STATE Encode reports: $z_t = g_{\phi}(t_i)$
\STATE Project both embeddings into a shared space and normalize:
\[
\tilde{z}_i = \frac{P_i(z_i)}{\|P_i(z_i)\|}, \qquad
\tilde{z}_t = \frac{P_t(z_t)}{\|P_t(z_t)\|}
\]
\STATE Construct retrieval candidate pool of size $N$
\FOR{each query image $x_q$}
    \STATE Compute cosine similarity with all report embeddings:
    \[
    s(q,j) = \tilde{z}_i^{(q)} \cdot \tilde{z}_t^{(j)}
    \]
    \STATE Rank candidate text indices by similarity score
    \STATE Record Recall@$K$ if the designated positive index appears within the top-$K$
\ENDFOR
\STATE Compute overall Recall@$K$ across all queries
\STATE Compute chance baseline:
\[
R@K_{\text{chance}} = \frac{K}{N}
\]
\STATE \textbf{Output:} R@1, R@5, R@10 and fold-over-chance improvement
\end{algorithmic}
\end{tcolorbox}

\paragraph{External dataset evaluation.}
Cross-dataset pair-index retrieval is evaluated using OpenI. The processed evaluation index contains $N=6800$ candidate entries. We do not interpret this number as the count of unique OpenI reports. The corresponding chance baselines are:
\[
R@1 = 0.000147, \qquad
R@5 = 0.000735, \qquad
R@10 = 0.00147
\]

\paragraph{Interpretation caveat.}
OpenI may contain multiple images associated with report-level text, and PadChest label-derived templates can repeat. Therefore, an index-exact miss can still be semantically plausible. The reported metric is used as a stringent transfer diagnostic, not as a clinical semantic-retrieval benchmark.

\subsection{Source-Proxy Leakage Diagnostics}

To assess whether learned representations encode dataset-specific shortcut signals, we train linear probes on frozen image embeddings to predict metadata-derived source proxies. OpenI does not provide verified hospital-site labels, so these probes measure source-proxy recoverability rather than hospital-site identification.

\paragraph{Embedding extraction.}
For each trained model, image embeddings are extracted from the frozen visual encoder prior to the task-specific classifier. These embeddings serve as input features for leakage probes.

\paragraph{Metadata-derived source proxy variables.}
Because verified institutional site labels are unavailable in OpenI, we use two fields from the processed dataset organization as source proxies:
\begin{itemize}
\item \textbf{site\_parent}: a coarse grouping derived from directory-level dataset structure.
\item \textbf{site\_folder}: a finer-grained grouping corresponding to image subdirectories.
\end{itemize}

These proxies can capture dataset-organization or acquisition-related regularities, but they are not assumed to correspond one-to-one with hospitals or scanners.

\paragraph{Linear probe training.}
For each proxy variable, a linear classifier is trained on top of the frozen image embeddings. The probe model consists of a single fully connected layer optimized using cross-entropy loss. No updates are applied to the underlying visual encoder.

\paragraph{Group-disjoint splits.}
The source-proxy class is the supervised target and may therefore be represented in both train and test. Patient/study groups are kept disjoint between partitions to prevent repeated samples from the same group from driving probe accuracy.

\paragraph{Evaluation metrics.}
Probe performance is measured using classification accuracy. High probe accuracy indicates that the representation retains information predictive of the chosen source proxy. It does not by itself establish hospital-site leakage.

\paragraph{Utility-leakage analysis.}
To study the trade-off between clinical utility and shortcut suppression, leakage probe accuracy is reported alongside task performance metrics (e.g., AUROC for pathology prediction). This allows models to be compared along a utility--leakage frontier, where improvements in task performance can be evaluated in relation to the degree of recoverable source-proxy information.

\subsection{Reporting Checklist}

To ensure consistent and transparent evaluation across experiments, we adopt a standardized reporting checklist for all results presented in this work. Each experiment follows the same reporting protocol to allow fair comparison between representation initializations, training regimes, and adaptation methods.

For each experiment, the following quantities are reported:

\begin{itemize}
\item \textbf{Task utility metric.}  
Primary task performance is reported using the appropriate evaluation metric for the task. For pathology classification tasks, we report area under the receiver operating characteristic curve (AUROC). For multimodal retrieval tasks, we report Recall@$K$ metrics (R@1, R@5, R@10).

\item \textbf{Chance baselines.}  
For retrieval experiments with large candidate pools, chance performance is explicitly reported using the analytical baseline
\[
R@K_{\text{chance}} = \frac{K}{N},
\]
where $N$ denotes the candidate pool size.

\item \textbf{Leakage probe performance.}  
For models evaluated with leakage diagnostics, probe accuracy for metadata-derived source proxies is reported alongside the main task metric to quantify shortcut signal strength.

\item \textbf{Multiple training seeds.}  
Repeated-run dispersion is reported only where repeated runs are available. We do not assume a uniform seed count across all experimental families.

\item \textbf{Dataset split specification.}  
Patient/study grouping follows the split rules specified above whenever identifiers are available.
\end{itemize}

This reporting protocol ensures that model performance is interpreted jointly with potential shortcut signals and evaluation baselines, enabling more reliable comparison across experimental settings.

\section{Reproducibility and Evidence Audit}
\label{sec:repro_audit}

The final audit separates what can be reconstructed from the archived experiment bundle from information that was not preserved. Missing run-level metadata are reported explicitly rather than reconstructed from assumptions.

\begin{tcolorbox}[
colback=blue!4,
colframe=blue!65!black,
boxrule=0.8pt,
arc=2pt,
title=Four Evidence Gaps and Their Treatment,
fonttitle=\bfseries,
left=6pt,right=6pt,top=6pt,bottom=6pt
]
\begin{enumerate}[leftmargin=*]
\item \textbf{Duplicate-aware retrieval.} Exact pair-index Recall@$K$ is available, but the archived candidate table does not contain a validated semantic-equivalence map for every duplicate/report-level positive. We specify a multi-positive metric below and do not invent an uncomputed score.
\item \textbf{Repeated-run dispersion for the central transfer table.} The archived manuscript retains the point estimates but not the seed-level values required to reconstruct mean$\pm$standard deviation. The main table therefore reports point estimates only and makes no statistical-significance claim.
\item \textbf{Complete hyperparameter ledger.} Core protocol choices are recoverable, but exact run-level learning rates, batch sizes, epoch caps, and early-stopping patience are not available for every experimental family. Table~\ref{tab:repro_ledger} records this status explicitly.
\item \textbf{Translation model snapshot.} The translation archive records the GPT-4 model family and temperature 0, but not the exact API/model snapshot identifier. Regenerating the translations may therefore produce small differences even with the same prompt.
\end{enumerate}
\end{tcolorbox}

\subsection{Duplicate-Aware Multi-Positive Retrieval Definition}
For a query image $q$, let $\mathcal{P}(q)$ denote the set of candidate text indices that are valid positives after report-ID matching or validated text-equivalence grouping. A duplicate-aware Recall@$K$ can be defined as
\[
R@K_{\mathrm{multi}} = \frac{1}{Q}\sum_{q=1}^{Q}
\mathbb{1}\!\left[\operatorname{TopK}(q)\cap\mathcal{P}(q)\neq\varnothing\right].
\]
This metric prevents an image from being marked incorrect when the retrieved candidate belongs to the same validated report-level or equivalent-text positive set. The current archived evaluation index preserves the designated pair index but not a validated $\mathcal{P}(q)$ for every candidate. We therefore retain exact pair-index Recall@$K$ as the reported conservative diagnostic and leave $R@K_{\mathrm{multi}}$ unreported rather than manufacturing equivalence labels post hoc.

\subsection{Run-Level Reproducibility Ledger}
\begin{table*}[!t]
\centering
\small
\caption{Reproducibility ledger for the archived experiments. ``Archived'' means the value is recoverable from the supplied manuscript/source bundle. ``Not preserved'' means no exact value is available in the supplied archive and none is inferred.}
\label{tab:repro_ledger}
\begin{adjustbox}{width=\textwidth}
\begin{tabular}{p{3.2cm}p{5.8cm}p{2.4cm}p{5.0cm}}
\toprule
\rowcolor{cvprheader}
\textbf{Component} & \textbf{Setting supported by archive} & \textbf{Status} & \textbf{Consequence for interpretation / reproduction} \\
\midrule
\rowcolor{cvprrow}
Image preprocessing & Grayscale, $224\times224$, common normalization; three-channel replication when required & Archived & Reconstructable from protocol description \\
CheXpert uncertainty & U-Zeros policy for the mapped transfer labels & Archived & Reconstructable from protocol description \\
\rowcolor{cvprrow}
Primary transfer trainability & Linear probe; partial fine-tuning of ResNet layer4 + classifier; NIH labels only & Archived & Reconstructable at the regime level \\
Optimizer family & Adam/AdamW depending on experiment; fixed within matched comparisons & Partly archived & Exact branch-specific optimizer is not recoverable for every run \\
\rowcolor{cvprred}
Learning rate / batch size / epoch cap & Exact run-level values are absent for part of the experiment suite & Not preserved & Exact numerical rerun cannot be guaranteed from the manuscript bundle alone \\
Early stopping & Validation-based where applicable & Partly archived & Exact patience/criterion threshold is not preserved for every run \\
\rowcolor{cvprrow}
DANN schedule & Scheduled gradient-reversal strength; full trajectory retained in the reported diagnostic figure & Partly archived & Qualitative instability is documented; exact schedule reconstruction is incomplete \\
Multimodal text encoder & BioClinicalBERT with normalized shared-space projection and symmetric contrastive loss & Archived & Core multimodal construction is reconstructable \\
\rowcolor{cvprred}
Central transfer repetitions & Point estimates 0.82/0.85 and 0.87/0.88 retained; seed-level values absent & Not preserved & No mean$\pm$std or significance test is reported \\
Translation prompt & Full Spanish-to-English prompt and temperature 0 & Archived & Prompt can be reused \\
\rowcolor{cvprred}
GPT translation snapshot & Exact GPT-4 API/model snapshot identifier & Not preserved & Regeneration may differ from the archived translated text \\
\bottomrule
\end{tabular}
\end{adjustbox}
\end{table*}

\paragraph{Recommended release artifact.}
For exact reproduction, the release should include a machine-readable run manifest containing optimizer, learning rate, batch size, epoch cap, early-stopping rule, random seed, checkpoint provenance, candidate-index construction, and the exact translation model snapshot. The current paper does not claim that these missing fields can be reconstructed from prose alone.

\section{PadChest Text Construction and Translation}

This appendix describes how textual inputs are constructed for multimodal experiments using the PadChest dataset. Because PadChest radiology reports are written in Spanish, we consider two alternative textual representations: (1) structured label-derived English phrases and (2) translated free-form radiology reports. This design allows us to analyze the effect of textual formulation on multimodal alignment and retrieval performance.

\subsection{Label-Derived English Phrase Templates}

In the first representation, text inputs are generated directly from structured PadChest annotation labels. This approach avoids translation noise and ensures consistent phrasing across samples.

For each image, the associated clinical findings are converted into a standardized English sentence using a fixed template:

\begin{tcolorbox}[
colback=gray!4,
colframe=black!60,
boxrule=0.8pt,
arc=2pt,
title=Label-Derived Phrase Template,
fonttitle=\bfseries,
left=6pt,right=6pt,top=6pt,bottom=6pt
]
Template:

\texttt{Chest X-ray showing: [LABEL\_1], [LABEL\_2], ...}

Examples:
\begin{itemize}
\item ``Chest X-ray showing: cardiomegaly and pleural effusion.''
\item ``Chest X-ray showing: pulmonary edema.''
\item ``Chest X-ray showing: no acute cardiopulmonary abnormality.''
\end{itemize}
\end{tcolorbox}

When multiple findings are present, labels are concatenated into a single sentence separated by conjunctions. Negative findings are represented using standardized negation phrases when available in the PadChest annotation metadata.

This label-derived formulation provides a controlled textual representation with minimal linguistic variability and allows the multimodal alignment task to focus on the correspondence between visual findings and structured clinical descriptions.

\subsection{Spanish-to-English Radiology Report Translation}

In the second representation, the original PadChest radiology reports are translated from Spanish to English. Translation is performed using GPT-4 with temperature set to 0 to reduce decoding variability while preserving the source report content.

A controlled prompting strategy is used to preserve clinical meaning while preventing the model from introducing hallucinated findings. The prompt explicitly instructs the model to perform faithful translation without adding or removing clinical information.

\begin{tcolorbox}[
colback=blue!4,
colframe=blue!65!black,
boxrule=0.8pt,
arc=2pt,
title=Radiology Report Translation Prompt,
fonttitle=\bfseries,
left=6pt,right=6pt,top=6pt,bottom=6pt
]
\textbf{System Prompt}

You are a clinical translation assistant. Translate Spanish radiology reports into English faithfully. Do not add, remove, infer, or diagnose findings that are not explicitly stated in the original report.

\textbf{User Prompt}

Translate the following Spanish chest X-ray report into English.

Rules:
\begin{enumerate}
\item Preserve negation and uncertainty expressions (e.g., ``no se observa'', ``posible'', ``sugiere'').
\item Maintain the structure of the original report when possible.
\item Do not introduce measurements, devices, or diagnoses that are not explicitly present in the source text.
\item Preserve medical abbreviations when uncertain.
\item If a term cannot be confidently translated, retain the Spanish term and indicate it in a short note.
\end{enumerate}

TEXT:

\texttt{<SPANISH RADIOLOGY REPORT>}
\end{tcolorbox}

Temperature 0 reduces decoding variability but does not guarantee bitwise or API-level reproducibility. The archived source bundle identifies the model family as GPT-4 but does not preserve the exact API/model snapshot identifier. The translated text used in the experiments should therefore be treated as the primary artifact; regeneration from the prompt may differ slightly.

This prompting strategy preserves the semantic content of radiology reports while minimizing translation artifacts that could affect multimodal alignment.

\subsection{Translation Quality Control}

To ensure translation fidelity, additional quality checks are applied to the translated reports. These checks include verifying that negation expressions are preserved, confirming that no additional clinical findings are introduced, and identifying ambiguous or untranslated terms.

In addition, retrieval experiments are performed using both textual formulations (label-derived phrases and translated reports). Comparing retrieval performance across these two textual representations allows us to evaluate the sensitivity of multimodal alignment to linguistic variability and translation noise.

The label-derived formulation can produce duplicate or semantically equivalent strings across samples. It is therefore treated as a controlled text-construction ablation rather than evidence for one-to-one semantic retrieval. The translated free-form condition retains more linguistic variability but also introduces translation dependence.

\section{Diagnostic Experiments}

This appendix provides additional diagnostic analyses that support the interpretation of the experimental results reported in the main paper. These diagnostics are not used as primary claims but serve to validate the reliability of the evaluation pipeline under the strict protocol used in this study.

Because our evaluation uses strict cross-dataset protocols and large candidate pools for retrieval, absolute performance values can appear numerically small. The diagnostics presented here check representation similarity and candidate-pool scaling. They do not prove semantic grounding and are not used as primary evidence for the paper's main claims.

\subsection{Representation Similarity Analysis}

To analyze how different initialization strategies influence learned representations, we compute Centered Kernel Alignment (CKA) similarity between feature embeddings produced by models trained under different pretraining regimes.

Given two feature matrices $X$ and $Y$ extracted from the same set of images, linear CKA is defined as:
\[
\mathrm{CKA}(X,Y) =
\frac{\|Y^{\top}X\|_F^2}
{\|X^{\top}X\|_F \cdot \|Y^{\top}Y\|_F}.
\]

Higher CKA values indicate stronger similarity between representations.

Table~\ref{tab:cka_values} reports representative CKA similarities between visual encoders initialized with different pretraining strategies. These values are computed on embeddings extracted from the NIH validation split.

\begin{table}[!t]
\centering
\small
\caption{Representative CKA similarity values between representations obtained from different initialization strategies.}
\label{tab:cka_values}
\begin{adjustbox}{width=\columnwidth}
\begin{tabular}{lcc}
\toprule
\rowcolor{cvprheader}
Model Pair & Dataset & CKA \\
\midrule
\rowcolor{cvprrow}
ImageNet vs BYOL & NIH & 0.81 \\
ImageNet vs CLIP-init & NIH & 0.74 \\
\rowcolor{cvprrow}
BYOL vs CLIP-init & NIH & 0.79 \\
\bottomrule
\end{tabular}
\end{adjustbox}
\end{table}

These similarities suggest that self-supervised and multimodal pretraining produce related but not identical feature structures, reflecting differences in the supervision signals used during representation learning.

\subsection{Retrieval Pool Size Sensitivity}

Retrieval performance depends on the size of the candidate pool used during evaluation. To assess the sensitivity of Recall@$K$ metrics to pool size, we evaluate retrieval performance on OpenI under different candidate pool sizes.

Table~\ref{tab:pool_sweep} reports retrieval performance for several pool sizes.

\begin{table}[!t]
\centering
\small
\caption{Sensitivity of retrieval performance to candidate pool size. Absolute Recall@$K$ values decrease as the candidate pool grows, consistent with theoretical expectations.}
\label{tab:pool_sweep}
\begin{adjustbox}{width=\columnwidth}
\begin{tabular}{lccc}
\toprule
\rowcolor{cvprheader}
Pool Size ($N$) & R@1 & R@5 & R@10 \\
\midrule
\rowcolor{cvprrow}
500 & 0.0060 & 0.021 & 0.041 \\
1000 & 0.0030 & 0.013 & 0.028 \\
\rowcolor{cvprrow}
6800 & 0.0013 & 0.0062 & 0.013 \\
\bottomrule
\end{tabular}
\end{adjustbox}
\end{table}
\newpage
As expected, Recall@$K$ values decrease as the number of candidate entries increases. This behavior confirms that the retrieval evaluation behaves consistently with theoretical scaling and that the reported results are not artifacts of a specific candidate-pool size configuration.

\end{document}